\documentclass{article}

\PassOptionsToPackage{numbers,sort,compress}{natbib}

\usepackage[preprint]{neurips_2023}

\usepackage{graphicx} 
\usepackage{float}    

\usepackage[utf8]{inputenc} 
\usepackage[T1]{fontenc}    
\usepackage{lmodern}        
\usepackage{hyperref}       
\usepackage{url}            
\hypersetup{hidelinks}
\usepackage{booktabs}       
\usepackage{amsfonts}       
\usepackage{nicefrac}       
\usepackage{microtype}      
\usepackage{xcolor}         

\graphicspath{{images/}}

\title{Exploring Agentic Workflows for Generating High Quality Math Visual Aids}

\author{
  Rizwaan Malik \\
  Stanford Graduate School of Education \\
  \texttt{rizmalik@stanford.edu} \\
  \AND
  Ashna Khetan \\
  Department of Computer Science \\
  \texttt{ashnak@stanford.edu} \\
  \And
  Isabel Sieh \\
  Department of Computer Science \\
  \texttt{isabelrs@stanford.edu} \\
  \And
  Samin Khan \\
  Stanford Graduate School of Education \\ \texttt{samink@stanford.edu}
}

\begin{document}
\maketitle

\begin{abstract}
    Mathematical diagrams play a crucial role in K-12 education, both as problem components and as scaffolding for student comprehension \citep{Arcavi2003}. However, current AI tools, including Large Language Models (LLMs), struggle to reliably generate accurate and pedagogically sound visual diagrams, even when provided with detailed descriptions \citep{Malik2025, Roy2024}. While state-of-the-art LLMs achieve a 73.9\% success rate in generating diagrams for middle-school math problems \citep{Jain2023}, a significant gap remains. To address this, we introduce an agentic workflow that empowers LLM agents to evaluate the quality of generated visuals and use this feedback to dynamically improve their outputs in real time. This self-improvement loop aims to enhance the accuracy and educational appropriateness of AI-generated diagrams. Our research investigates the following questions: First, can LLMs accurately generate quality assurance (QA) questions for a visual aid, given specific criteria for visual aid quality? Second, given valid QA questions, can Vision Language Models (VLMs) effectively vet generated K-12 visual aids and use this to iteratively improve outputs? We conduct an exploratory evaluation of our agentic workflow. We identify key areas for improvement, including enhancing spatial reasoning and ensuring comprehensive coverage of diagram features in QA questions. Our results provide preliminary evidence that this approach can improve the reliability and educational value of AI-generated mathematical diagrams.
\end{abstract}

\section{Introduction}

Mathematical diagrams are fundamental to K-12 education, serving both as essential components of problems and as scaffolding tools for student comprehension \citep{Arcavi2003}. Diagrams help to visualize abstract concepts, making them more accessible and understandable for students of all learning styles.  For example, diagrams can clarify geometric relationships, illustrate proportional reasoning, and support problem-solving strategies. The ability to effectively create and utilize mathematical diagrams is therefore a crucial skill for both students and educators. However, manually creating these diagrams can be time-consuming and require a certain level of technical expertise.

The advent of Artificial Intelligence (AI) offers a promising opportunity to automate the generation of mathematical diagrams, potentially freeing up educators' time and providing students with personalized learning experiences. AI tools show promise in supporting teachers with lesson preparation; however, they still struggle with reliably generating visual diagrams \citep{Belouadi2024, Mondal2024}. This is particularly true for diagrams that require precise geometric relationships or symbolic representations. Furthermore, generated diagrams must not only be accurate but also pedagogically sound, meaning they should effectively convey the intended mathematical concepts and be appropriate for the target student population.

While progress has been made in using Large Language Models (LLMs) to generate code or descriptions for mathematical diagrams, the accuracy of these generated visuals remains a significant challenge. When provided with specific, detailed diagram descriptions, state-of-the-art LLMs achieve a 73.9\% success rate within the middle-school domain \citep{Jain2023}. This indicates that even with precise instructions, current LLMs struggle to consistently generate diagrams that meet the required standards of accuracy and pedagogical effectiveness. Closing this gap is essential to unlock the full potential of AI in mathematics education.

To address the remaining gap in diagram accuracy, we introduce an agentic workflow aimed at having LLM agents evaluate generated visuals and use this feedback to improve outputs dynamically and in real time. Our approach leverages the strengths of both LLMs and Vision Language Models (VLMs) to create a self-improving system for generating mathematical diagrams. The workflow involves the generation of diagrams from textual descriptions, followed by the automated creation of quality assurance (QA) questions that assess the accuracy and clarity of the visual. A VLM then evaluates the diagram based on these QA questions, providing feedback that is used to iteratively refine the diagram. Our research focuses on the following key research questions:
\begin{enumerate}
    \item Given criteria for visual aid quality, can LLMs accurately generate quality assurance (QA) questions for a visual aid?
    \item Given valid QA questions, can VLMs effectively vet generated K-12 visual aids and use this to iteratively improve outputs?
\end{enumerate}

We hypothesize that this agentic workflow can improve the quality and accuracy of AI-generated mathematical diagrams. By enabling a continuous feedback loop, we aim to create a system that can learn from its mistakes and generate increasingly effective visual aids for mathematics education. We present our approach and experimental results as an initial step toward more reliable AI-assisted mathematics diagram generation.

\begin{figure}[H]
    \centering
    \includegraphics[width=0.2\textwidth]{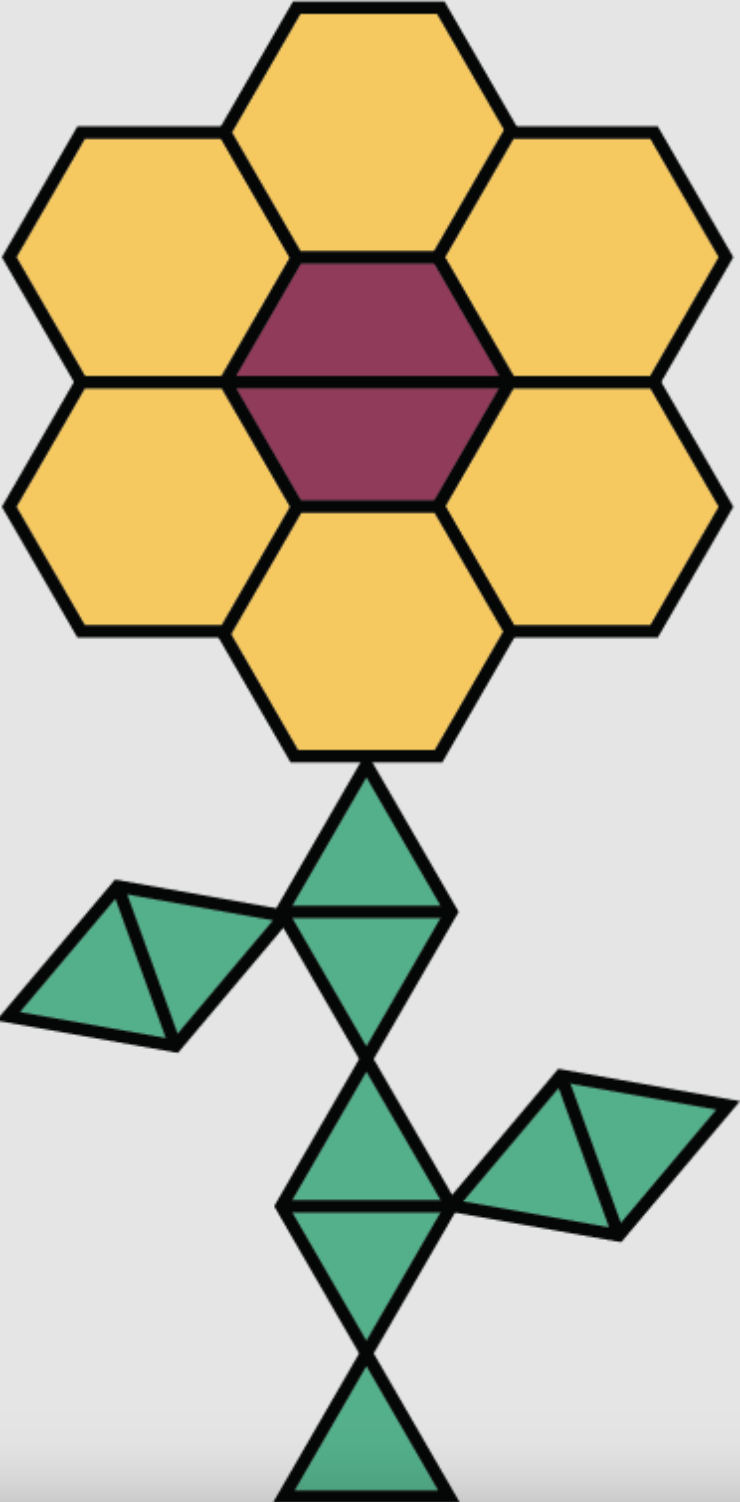}
    \caption{Sample benchmark diagram from Illustrative Mathematics. This is the associated student task statement: This flower is made up of yellow hexagons, red trapezoids, and green triangles. Write sentences to describe the ratios of the shapes that make up this pattern. How many of each shape would be in two copies of this flower pattern?}
    \label{fig:sample}
\end{figure}

\section{Related Work}
\subsection{Mathematical Visualization Generation}
The creation of accurate, pedagogically effective visual aids for mathematics remains a challenging problem requiring specialized approaches. VISTA (Visual Integrated System for Tailored Automation) introduces a multi-agent framework that uses LLMs to generate mathematical problem text together with aligned visualizations \citep{vista2024}. Its emphasis on consistency between problem statements and visual components is closely related to our goal of making diagram generation more reliable in educational settings.

Recent work in scientific diagram generation provides valuable technical foundations for automated visualization systems. \textit{AutomaTikZ} uses TikZ as an intermediate representation for text-guided synthesis of scientific vector graphics \citep{Belouadi2024}. While its focus on technical scientific illustrations differs from our educational emphasis, its combination of automatic and human evaluation offers a useful model for assessing visual fidelity.

MathVerse provides complementary evidence that multimodal LLMs can appear successful on visual math tasks without fully using the diagram, motivating more careful evaluation of diagram understanding rather than relying on final-answer correctness alone \citep{mathverse2024}.

\subsection{Evaluation Methodologies for Mathematical Visuals}
The evaluation of generated mathematical visuals presents unique challenges that recent research has begun to address. \textit{DrawEduMath} contributes an expert-annotated dataset of student-drawn math images paired with teacher-generated QA items \citep{drawe2024}. While its focus on handwritten solutions differs from our computer-generated focus, its combination of teacher-authored and synthetic QA materials provides a useful template for evaluating mathematical reasoning over images. Our work extends these principles to algorithmically generated educational visuals, particularly targeting the K-12 context where creative representations, such as floral diagrams, often supplement mathematical rigor.

These evaluation-focused works complement technical generation systems such as GlycoDraw, which demonstrates the value of domain-specific visualization standards for producing consistent figures \citep{glycodraw2023}. The integration of rigorous evaluation protocols with domain-aware generation systems suggests promising pathways for developing educational visualization tools that balance technical accuracy with pedagogical effectiveness \citep{glycodraw2023,Belouadi2024}.

\subsection{Agentic Workflows and LLM-based Systems}
The emergence of agentic workflows has transformed approaches to complex problem-solving across domains, with significant implications for mathematical visualization generation. The "Automated Design of Agentic Systems" (ADAS) framework presents a foundational approach where agents defined in code can be automatically discovered by meta-agents programming increasingly sophisticated variants \citep{adas2024}. This meta-agent search paradigm shows how structured agent design can improve performance across multiple domains, including mathematics.

A prototypical application of agentic approaches to visualization is demonstrated in work on automatic creation of P\&ID diagrams from natural language descriptions \citep{pid2024}. That system implements a multi-step workflow for structured, iterative diagram creation directly from natural language prompts. While focused on engineering diagrams, the principles transfer readily to mathematical visualization generation.

Data Interpreter represents an LLM-based agent designed specifically for end-to-end data science problems \citep{datainterpreter2024}. Its Hierarchical Graph Modeling approach decomposes complex problems into manageable subproblems with dynamic node generation, illustrating the value of structured decomposition for mathematical applications.

Polymind introduces a visual diagramming tool that leverages multiple LLM-powered agents through a parallel collaboration workflow rather than traditional turn-taking interactions \citep{polymind2025}. By defining discrete "microtasks" that simulate group collaboration scenarios, Polymind enables users to orchestrate multiple simultaneous processes rather than repetitively prompting a single chatbot. This orchestration approach offers valuable insights for coordinating multiple specialized agents in mathematical visualization generation.

The concept of Math Agents as a distinct category of AI systems is explored in research examining computational infrastructure and mathematical embedding \citep{mathagents2023}. The authors propose a GPT-based workflow to convert equations from literature into LaTeX and Python formats, suggesting a potential shift from "big data" to "big math" through specialized agents that interact with mathematical structures.

\section{Problem Statement}
The effective teaching of mathematics fundamentally relies on visual representations to make abstract concepts concrete and accessible. This is particularly crucial for struggling students who benefit from seeing multiple representations of mathematical ideas. While recent advances in AI have enabled the generation of differentiated text content and practice problems for math instruction, there remains a critical gap: the inability to automatically generate and customize high-quality visual aids that align with instructional needs.

We envision a two-stage pipeline for AI-assisted educational content creation. In the first stage, an LLM generates curriculum content and image descriptions. In the second stage, an agentic workflow transforms these image descriptions into appropriate visual aids. Our work focuses on this critical second stage, where the challenge lies in ensuring generated visuals effectively serve their intended pedagogical purpose.

A fundamental barrier to progress in this domain is the lack of robust evaluation mechanisms. The key question we address is: How can we systematically assess whether AI-generated visual aids accurately reflect their intended pedagogical purpose as specified in the image descriptions? This capability is essential not only for quality assurance but also for enabling self-improving agentic workflows that can learn from feedback and iteratively enhance their outputs.

Without reliable methods to evaluate the fidelity and educational appropriateness of AI-generated visual aids, the potential impact of AI in mathematics education remains constrained. Current evaluation approaches for image generation focus primarily on aesthetic quality and general semantic alignment, failing to capture the nuanced requirements of mathematical visualization and pedagogical effectiveness.

Our research addresses this gap by developing a comprehensive evaluation framework for assessing the alignment between intended mathematical concepts (as expressed in image descriptions) and their visual realizations. This framework will serve as a foundation for developing more sophisticated AI systems that can reliably generate and refine visual aids, ultimately supporting more effective and inclusive mathematics instruction.

\section{Methods}

\begin{figure}[H]
    \centering
    \includegraphics[width=0.8\textwidth]{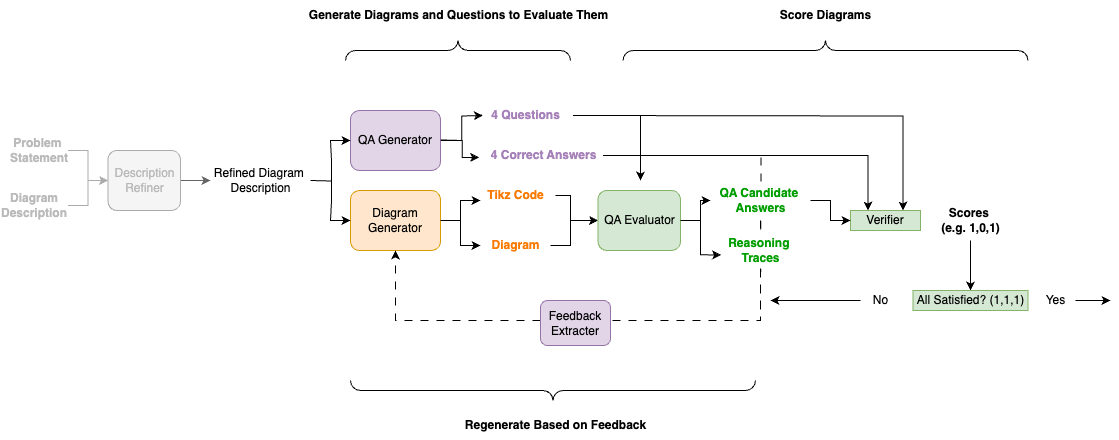}
    \caption{Workflow Diagram for Auto-Evaluation Process}
    \label{fig:workflow-diagram}
\end{figure}

We propose a self-improvement mathematical diagram generation pipeline as shown in Figure 2. Specifically, given an enhanced description of a mathematical diagram, we generate an initial diagram using TikZ and four open-ended QA question-answer pairs. We use these questions to assess the quality of the initial diagram, then feed the unsatisfied questions back into the generator, regenerating the diagram with this feedback. This refinement loop continues until either all questions are satisfied or the pipeline reaches a maximum number of cycles.

\subsection{QA Generator}
In our initial experimentation, we had the QA generator create yes/no questions about the quality of the diagram, where a yes would indicate that the metric was satisfactorily met (e.g., are there exactly 6 yellow regular hexagons, 2 red trapezoids, and 9 green equilateral triangles in the illustration?). Cohen's \(\kappa\) was 0.11, with 68\% raw agreement, which indicated slight agreement. Since open-ended questions are less likely to be answered correctly by chance \citep{Antol2015}, we opted to use open-ended questions. This requires adding a verifier model to the pipeline to determine whether the diagram satisfies the open-ended question. For example: "Describe the TikZ commands used to ensure the hanger bar appears horizontal and balanced. Does the code explicitly specify equal vertical distances from the support to each end of the bar? Why or why not?" The trade-off is that this approach relies on the quality of both the evaluator and the verifier to provide accurate diagram feedback. 

In addition, we experimented with the categories of questions. Although accuracy and clarity were consistent categories, we originally included an "appropriateness" category with questions such as: "Is this double number line diagram an appropriate visual aid for seventh-graders learning about percentages and their equivalent dollar values?" This question was almost always scored positively by both the evaluator and human annotators and did not provide valuable information. We replaced it with a miscellaneous question that checked qualities such as avoiding hints or excluding extra information (e.g., "Does the diagram avoid providing any hints or answers about the value of each individual box, ensuring the student must perform the calculation?"). More often than not, this question was also of little help and too vague to be determined by an LLM reliably, so we eventually settled on accuracy and clarity as the only categories. 

Our final QA generator creates four open-ended questions, two each from the categories of accuracy and clarity, along with their correct answers.

\subsection{VLMs and LLMs as a QA Evaluator}
Originally, we harnessed the VLM capabilities of the state-of-the-art models to evaluate generated diagrams as images, rather than as the code that creates them. However, we noticed that VLMs can be poor at analyzing complex diagrams, particularly the very attributes that the generators also tend to get wrong. Broadly, they often miscounted items or grid units and have limited spatial awareness.

This means that our evaluation loop does not always provide valuable feedback to the regeneration process, which weakens the purpose of the feedback loop. These signals, such as the number of elements and amount of space, are often easier to determine from the code that renders the image. These values may appear explicitly or be derived from the textual logic of the code, making them easier for an LLM to determine from code than for a VLM to infer from the image. Therefore, we switched to generating QA questions for, and evaluating these questions on, the TikZ code itself. In our final approach, we combine the two methods. We saw much higher human agreement with these questions and evaluated answers, indicating that the combined code-and-image representation supports a stronger evaluation model.

A final change that resulted in more accurate QA answers was adding chain-of-thought prompting to evaluation, which produced the highest performance gains when testing with yes/no questions and gains with open-ended questions as well.

\subsection{Full Pipeline}
In summary, our pipeline receives as input a refined description of a mathematical diagram. Our \textbf{Diagram Generator} creates an initial diagram, rendered from generated TikZ code, and our \textbf{QA Question Generator} creates a set of four QA question-answer pairs. Our \textbf{QA Evaluator} takes the code, image diagram, and four QA questions, then evaluates answers based on the diagram. Our \textbf{Verifier} compares these answers with the QA correct answer and determines whether each question's criteria were satisfied (1) or not (0). If all questions were answered correctly, the diagram is returned; otherwise, the unsatisfied question answers are transformed into feedback by our \textbf{Feedback Extractor}, and the diagrams are regenerated until all questions are satisfied or the maximum number of cycles is reached.

\section{Experiments}

We report an exploratory evaluation of the workflow, focusing on descriptive agreement rates, regeneration behavior, and qualitative error patterns.

\subsection{Evaluating the QA Evaluator}

To evaluate the QA Evaluator, we measured human agreement with the QA Evaluator's judgments. We considered three scenarios (where Evaluator considers both the human and the pipeline component):

\begin{itemize}
    \item \textbf{Diagram-Only:} Evaluators were provided with the generated diagram and QA questions.
    \item \textbf{TikZ Code-Only:} Evaluators were provided with the TikZ code and QA questions.
    \item \textbf{Code + Diagram:} Evaluators were provided with both the TikZ code and the diagram, along with QA questions.
\end{itemize}

The results, summarized in Table~\ref{tab:qa_eval}, show that the human agreement increased when both the code and diagram were available, demonstrating the importance of multi-modal information for verification. An example where evaluating code misses what image understanding captures is seen in the Appendix. 

\begin{table}[h!]
    \centering
    \caption{Human Agreement with QA Evaluator under different conditions.}
    \begin{tabular}{lc}
        \toprule
        Experiment & Human Agreement \\
        \midrule
        Diagram-Only & 68\% \\
        TikZ Code-Only & 81\% \\
        Code + Diagram & 97\% \\
        \bottomrule
    \end{tabular}
    \label{tab:qa_eval}
\end{table}

\subsection{Evaluating Regeneration}

We evaluated the impact of our regeneration loop on the quality of the generated diagrams. We asked three human annotators to score the generated diagram on a scale of 1-5 based on the diagram description and the context of usefulness for K-12 education. 

In our results, shown in Table~\ref{tab:regeneration}, we include the percentage of problems that are sent for another loop. Thirty percent of diagrams were considered complete after initial creation; however, after one regeneration, all remaining diagrams were still sent into another loop. This indicates that the issues in those diagrams are hard to fix with the feedback we provide. We also include the percentage of questions that the Evaluator marks as unsatisfactory, which decreases in every iteration. Human-perceived quality scores improve modestly after each regeneration iteration as well.

\begin{table}[h!]
    \centering
    \caption{Impact of Regeneration on Diagram Quality.}
    \begin{tabular}{lccc}
        \toprule
        Round & \% Problems Generated & \% QA Unsatisfied & Avg Human Score \\
        \midrule
        Initial & 100\% & 24.3\% & 3.4 \\
        Regeneration 1 & 70\% & 18.3\% & 3.65 \\
        Regeneration 2 & 70\% & 15.9\% & 3.7 \\
        \bottomrule
    \end{tabular}
    \label{tab:regeneration}
\end{table}

\subsection{Error Analysis: QA Questions}
We conducted an error analysis to understand the limitations of the generated QA questions and VLM answers. While the agentic workflow showed promise, errors arose primarily from two key areas: (1) challenges in ensuring comprehensive coverage of all diagram features in the QA questions, and (2) limitations in the VLM's ability to accurately assess spatial relationships and object properties within the diagrams.

To further investigate these issues, we analyzed specific examples where the AI failed to provide accurate answers, highlighting the underlying reasons for these failures.
\begin{enumerate}
    \item \textbf{Insufficient Coverage of QA Questions:}  The generated QA questions sometimes failed to address all critical aspects of the diagram, leading to incomplete evaluations.  For instance, while a QA question might focus on the horizontal alignment of elements, it could overlook the accuracy of vertical spacing or the proportionality of different components.
    \item \textbf{Limitations in Spatial Reasoning and Object Property Evaluation:} VLM encountered difficulties in accurately evaluating specific aspects of diagrams:
    \begin{itemize}
        \item \textbf{Item/Unit Counts:} Accurately counting and verifying the number of items or units in a diagram proved challenging for the VLM.
        \item \textbf{Spatial Relationships:} Assessing spatial relationships between objects, such as relative positions, distances, and overlap, presented another hurdle.
    \end{itemize}
\end{enumerate}

\textbf{Example 1: Balanced Hanger Diagram Error}

Consider the example of a balanced hanger diagram.  The intended goal was to accurately depict a balanced hanger model.
\begin{figure}[H]
    \centering
    \includegraphics[width=0.3\textwidth]{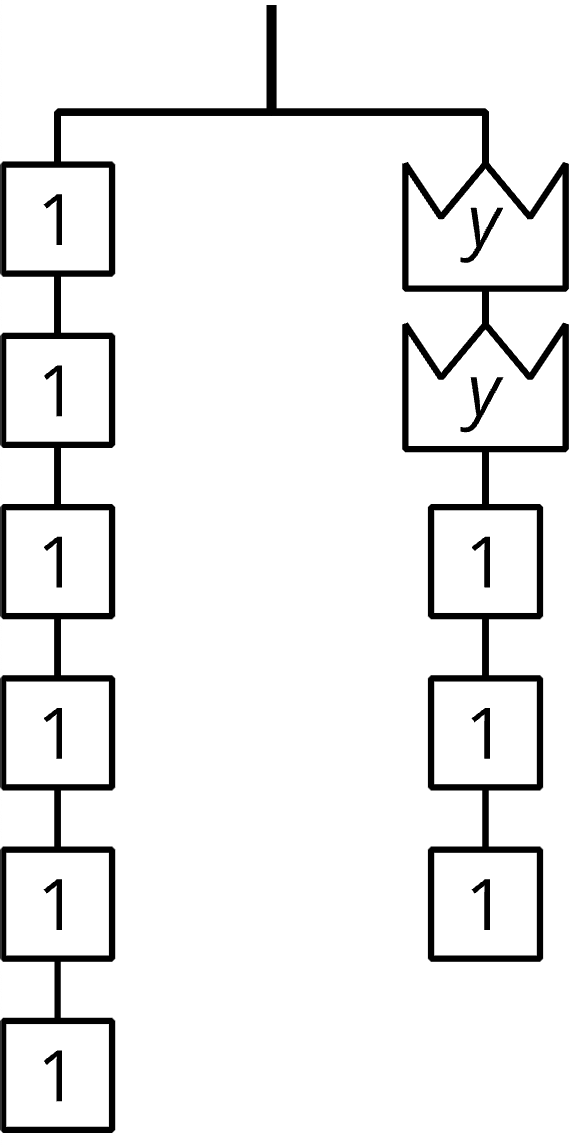}
    \caption{Balanced Hanger (Gold Standard)}
    \label{fig:balanced_golden}
\end{figure}

\begin{figure}[H]
    \centering
    \includegraphics[width=0.8\textwidth]{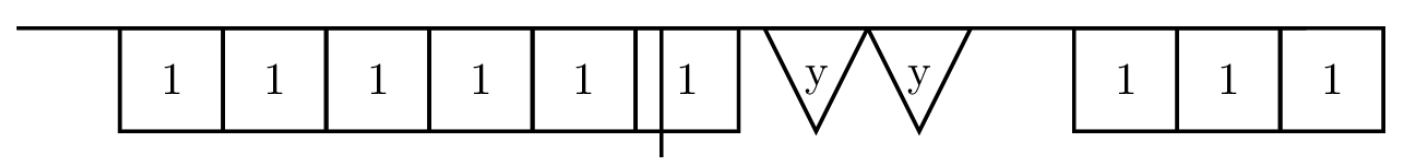}
    \caption{Initial AI Generated Diagram of Balanced Hanger}
    \label{fig:balanced_ai_1}
\end{figure}

\begin{figure}[H]
    \centering
    \includegraphics[width=0.8\textwidth]{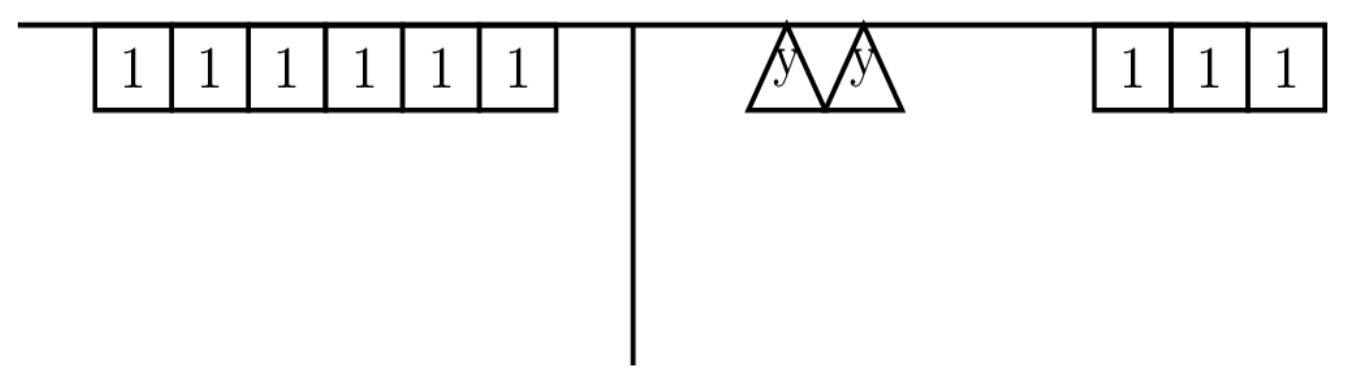}
    \caption{Regenerated AI Diagram of Balanced Hanger}
    \label{fig:balanced_ai_2}
\end{figure}

\begin{itemize}
    \item \textbf{Gold Standard Diagram:} The gold standard (Figure \ref{fig:balanced_golden}) shows the ideal diagram, representing the structure of a balanced hanger.
    \item \textbf{Initial AI-Generated Diagram:} As shown in Figure \ref{fig:balanced_ai_1}, the initial diagram depicts a balanced hanger with boxes hanging from it.
     \item \textbf{Regenerated AI Diagram:} As shown in Figure \ref{fig:balanced_ai_2}, even after regeneration, the boxes are not evenly spaced along the bar, failing to match the configuration of 
     Figure \ref{fig:balanced_golden}.
     \item \textbf{QA Question} "Describe the TikZ commands used to ensure the hanger bar appears horizontal and balanced. Does the code explicitly specify equal vertical distances from the support to each end of the bar?"
     \item \textbf{Gold Standard Answer (if code correct)} "The code should define \ldots{} such that the y-coordinate is constant across its length. There should be symmetry in y-values \ldots{}."
    \item \textbf{LLM Answer} "No; the horizontal bar is drawn from (-4,0) to (4,0) \ldots{} however, the box at (-3,0) overlaps with the vertical bar \ldots{}."
    \item \textbf{LLM Feedback for Follow-Up Regeneration} "The diagram's code should use node anchors with \texttt{inner sep=0pt} to center labels and avoid overlap with borders."
\end{itemize}

Even after the regeneration step, the diagram is still incorrect. The boxes are not evenly spaced along the bar, failing to match the balanced configuration of the gold standard diagram. The QA process could not effectively correct this issue. One potential reason is that the QA model missed the error, or that the feedback did not identify the specific failure mode, namely the uneven spacing of the boxes. Even with the regeneration step, the generated output diagram could not reach the balanced hanger standard due to a limitation in either the generated QA question or the accuracy of the QA model. In this case, the VLM correctly identifies the unbalanced state as incorrect but is unable to recover to a balanced diagram. This indicates a failure in the code synthesis step, not only the evaluative step.

\textbf{Example 2: L-shaped Polygon Diagram Error}
\begin{figure}[H]
    \centering
    \includegraphics[width=0.5\textwidth]{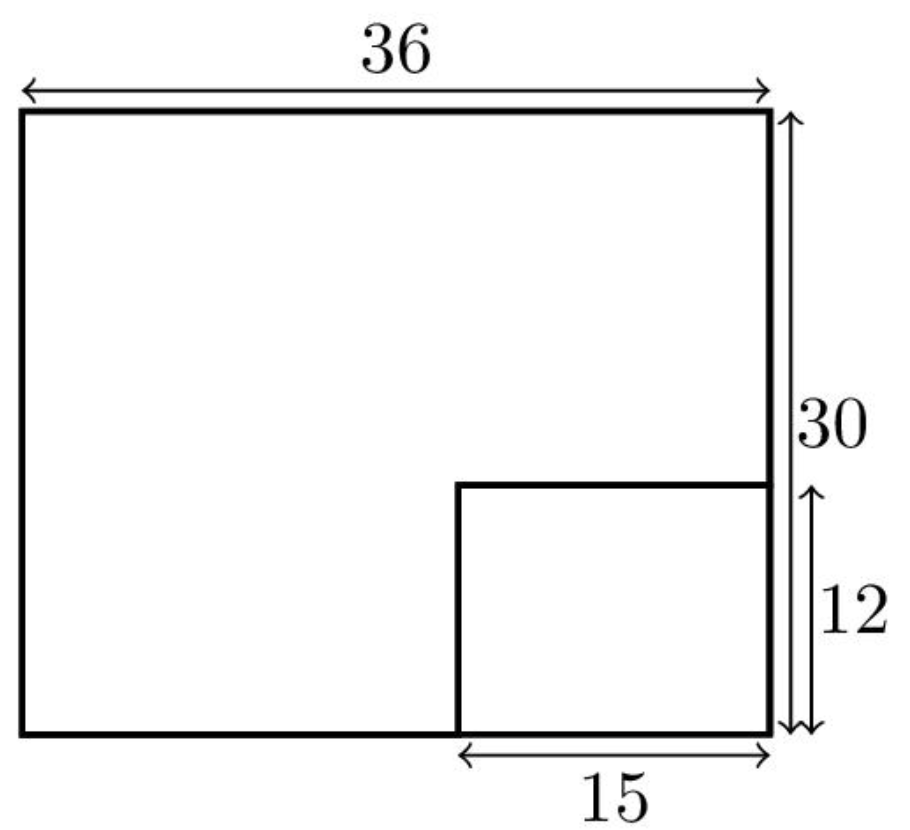}
    \caption{AI Generated Diagram of L-shaped Polygon}
    \label{fig:polygon_ai}
\end{figure}

\begin{figure}[H]
    \centering
    \includegraphics[width=0.5\textwidth]{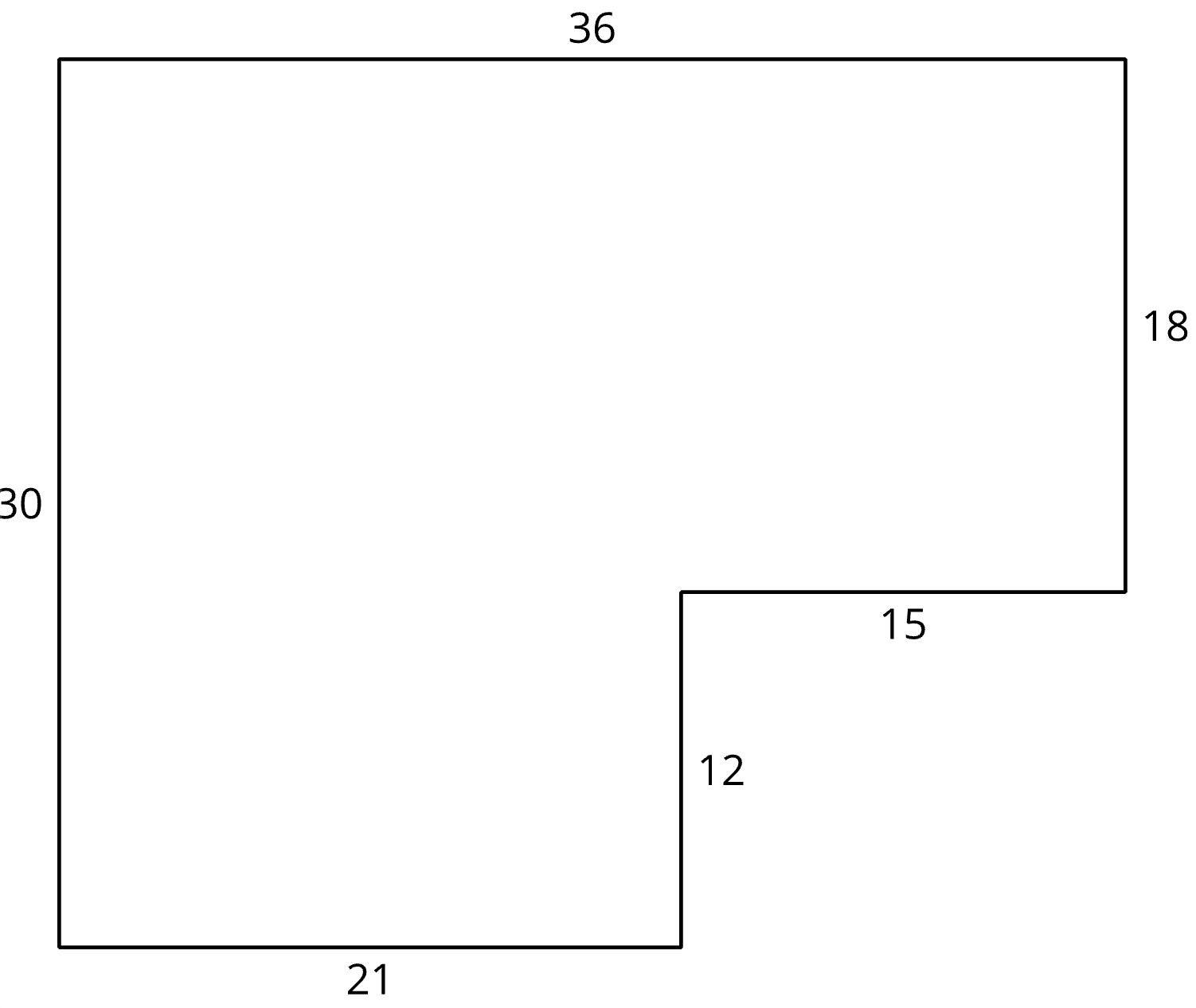}
    \caption{L-shaped Polygon (Gold Standard)}
    \label{fig:polygon_gold}
\end{figure}
\begin{itemize}
    \item \textbf{Gold Standard Diagram:} The gold standard (Figure \ref{fig:polygon_gold}) correctly depicts the polygon with accurate side length labels.
    \item \textbf{AI-Generated Diagram:} The initial AI-generated diagram (Figure \ref{fig:polygon_ai}) labels the polygon with the wrong measurements.
    \item \textbf{QA Question:} "How are the side lengths labeled in the TikZ code, and what ensures that they are placed clearly next to their corresponding sides without overlapping other elements or the polygon itself? Describe the method used for positioning the labels."
    \item \textbf{Incorrect AI Answer:} The AI's answer highlights the core problem: The side lengths are labeled with dimensions placed outside the rectangles using arrows; they ensure clarity by positioning labels using `node[midway, position]` such as `above`, `below`, and `right`, avoiding overlaps.
\end{itemize}

The error stemmed from limitations in the generated QA questions, which led the AI model to fail to recover the correct diagram. The generated QA focused on whether the positioning commands in the TikZ code were implemented correctly, but it did not ask whether the dimensions in the code matched the dimensions specified in the prompt. Furthermore, the generated AI answer failed to identify whether the dimensions provided in the TikZ code were the right dimensions for the original diagram. The dimensions themselves were incorrect, as can be seen by comparing against the gold standard. In this instance, the QA failed to ask the right questions to determine where the AI diagram was failing, so the model continued to state that the diagram was correct even though it reported factually wrong measurements.

Because this study originated as an exploratory course project, some experimental metadata, including exact model configurations, prompt settings, and evaluation sample counts, were not preserved. The reported results should therefore be interpreted as preliminary descriptive findings rather than a fully reproducible benchmark.

\section{Conclusion and Discussion}
In this work, we presented an agentic workflow for generating high-quality mathematical diagrams, addressing the critical need for customizable and pedagogically effective visual aids in mathematics education. Our approach leverages LLMs and a QA-based self-improvement loop to refine diagrams iteratively based on feedback.

Our experiments provide modest evidence for the promise of this approach, while also highlighting important limitations. Specifically, we found that:
\begin{itemize}
    \item Providing both the TikZ code and diagram greatly increases human agreement about the correctness of a QA answer, compared with providing either representation alone.
    \item Open-ended questions are more reliable than yes/no questions when evaluating diagram attributes with a VLM.
    \item The regeneration process yields modest improvements in human quality evaluation, from 3.4 initially to 3.7 after two regeneration rounds, but unresolved diagrams often still require additional regeneration.
\end{itemize}



This work opens up several avenues for future research and development:
\begin{itemize}
    \item \textbf{Improving QA Generation:} Develop more sophisticated strategies for generating QA questions that provide complete coverage of all relevant diagram features. This could involve using LLMs to automatically identify key aspects of the diagram and generate targeted questions for each, or even moving past questions to a more holistic criteria set.
    \item \textbf{Enhancing Spatial Reasoning:} Explore techniques for improving the spatial reasoning capabilities of VLMs, such as incorporating geometric priors or training on specialized datasets. Alternatively, explore non-vision approaches for evaluating spatial relationships based on the diagram's code representation. Either way, a fine-tuning or RAG-based approach using previous examples for diagram generation may also be promising on the spatial generation side.
    \item \textbf{Understanding Diagram Generation:} Further work is needed to provide more actionable feedback to the regeneration process. Additional experimentation could reveal which forms of feedback produce the largest changes in TikZ code generation.
\end{itemize}

Our research provides a promising step toward the automated generation of high-quality mathematical visual aids. By combining LLMs with a QA-based self-improvement loop, we can begin to create systems that generate customizable and pedagogically effective diagrams for mathematics education. The current results are encouraging but preliminary, and they show that future systems will need stronger QA coverage and more reliable regeneration before these methods can be deployed broadly.

\section{Appendix}

\subsection{TikZ Code and Visual Diagrams}
When code was evaluated for the following diagram, it appeared that the diagram had all six hexagons, but upon viewing the image, a VLM could tell that there effectively was one. This is because the hexagons were overlapping in the TikZ code.

\begin{verbatim}
% Hexagon coordinates
\foreach \i in {0, 60, 120, 180, 240, 300} {
    \begin{scope}[rotate=\i]
        \fill[hexagonColor] (0:1) -- (60:1) -- (120:1) -- (180:1) 
        -- (240:1) -- (300:1) -- cycle;
        \draw (0:1) -- (60:1) -- (120:1) -- (180:1) -- (240:1) 
        -- (300:1) -- cycle;
    \end{scope}
}
\end{verbatim}

\begin{figure}[h]
  \centering
    \includegraphics[width=200px]{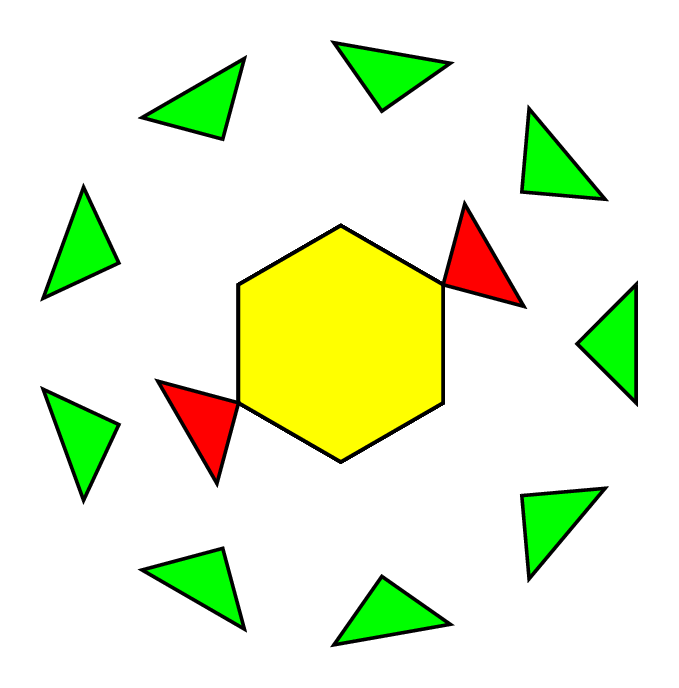}
    \caption{Initial Generation: 1 hexagon}
\end{figure}

\begin{figure}[h]
  \centering
    \includegraphics[width=200px]{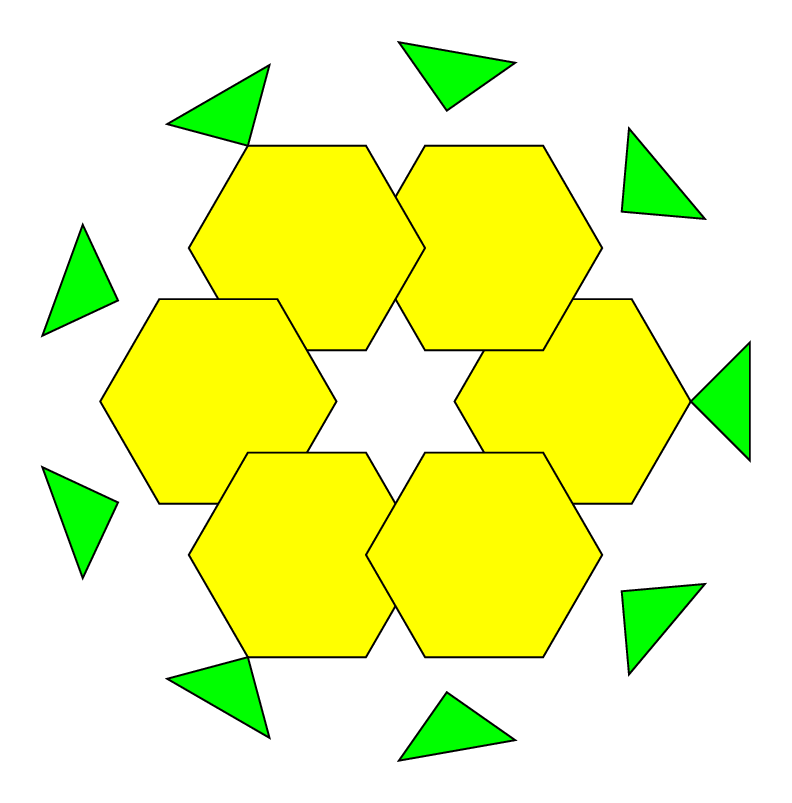}
    \caption{Updated Generation: 6 hexagons, 0 trapezoids}
\end{figure}

\subsection{Code}
All code can be found at \href{https://github.com/ashnakhetan/self-improve-math-diagrams}{https://github.com/ashnakhetan/self-improve-math-diagrams}

\begin{thebibliography}{99}
\bibitem{Arcavi2003}
Abraham Arcavi. 2003. The role of visual representations in the learning of mathematics. \textit{Educational Studies in Mathematics} 52, 3 (2003), 215--241. \url{https://doi.org/10.1023/A:1023284510913}

\bibitem{Antol2015}
Stanislaw Antol, Aishwarya Agrawal, Jiasen Lu, Margaret Mitchell, Dhruv Batra, C. Lawrence Zitnick, and Devi Parikh. 2015. VQA: Visual question answering. In \textit{Proceedings of the IEEE International Conference on Computer Vision}. 2425--2433. \url{https://doi.org/10.1109/ICCV.2015.279}

\bibitem{Malik2025}
Rizwaan Malik, Dorna Abdi, Rose Wang, and Dorottya Demszky. 2025. Scaffolding middle-school mathematics curricula with large language models. \textit{British Journal of Educational Technology} 56, 3 (2025), 1363--1386. \url{https://doi.org/10.1111/bjet.13571}

\bibitem{Roy2024}
Palak Roy, Helen Poet, Ruth Staunton, Katherine Aston, and David Thomas. 2024. \textit{ChatGPT in Lesson Preparation: A Teacher Choices Trial}. Education Endowment Foundation and National Foundation for Educational Research. \url{https://www.nfer.ac.uk/publications/chatgpt-in-lesson-preparation-a-teacher-choices-trial/}

\bibitem{Jain2023}
Rijul Jain, Wode Ni, and Joshua Sunshine. 2023. Generating domain-specific programs for diagram authoring with large language models. In \textit{Companion Proceedings of the 2023 ACM SIGPLAN International Conference on Systems, Programming, Languages, and Applications: Software for Humanity}. ACM, 70--71. \url{https://doi.org/10.1145/3618305.3623612}

\bibitem{Belouadi2024}
Jonas Belouadi, Anne Lauscher, and Steffen Eger. 2024. AutomaTikZ: Text-guided synthesis of scientific vector graphics with TikZ. In \textit{Proceedings of the Twelfth International Conference on Learning Representations}. \url{https://arxiv.org/abs/2310.00367}

\bibitem{Mondal2024}
Ishani Mondal, Zongxia Li, Yufang Hou, Anandhavelu Natarajan, Aparna Garimella, Sambaran Bandyopadhyay, and Jordan Boyd-Graber. 2024. SciDoc2Diagrammer-MAF: Towards generation of scientific diagrams from documents guided by multi-aspect feedback refinement. In \textit{Findings of the Association for Computational Linguistics: EMNLP 2024}. \url{https://aclanthology.org/2024.findings-emnlp.780/}

\bibitem{vista2024}
Jeongwoo Lee, Kwangsuk Park, and Jihyeon Park. 2024. VISTA: Visual integrated system for tailored automation in math problem generation using LLM. \textit{arXiv:2411.05423}. \url{https://arxiv.org/abs/2411.05423}

\bibitem{mathverse2024}
Renrui Zhang, Dongzhi Jiang, Yichi Zhang, Haokun Lin, Ziyu Guo, Pengshuo Qiu, Aojun Zhou, Pan Lu, Kai-Wei Chang, Peng Gao, and Hongsheng Li. 2024. MathVerse: Does your multi-modal LLM truly see the diagrams in visual math problems? In \textit{European Conference on Computer Vision}. \url{https://arxiv.org/abs/2403.14624}

\bibitem{drawe2024}
Sami Baral, Li Lucy, Ryan Knight, Alice Ng, Luca Soldaini, Neil T. Heffernan, and Kyle Lo. 2025. DrawEduMath: Evaluating vision language models with expert-annotated students' hand-drawn math images. In \textit{Proceedings of the 2025 Conference of the North American Chapter of the Association for Computational Linguistics: Human Language Technologies}. \url{https://aclanthology.org/2025.naacl-long.352/}

\bibitem{glycodraw2023}
Jon Lundstrom, James Urban, Luc Thomes, and Daniel Bojar. 2023. GlycoDraw: A Python implementation for generating high-quality glycan figures. \textit{Glycobiology} 33, 11 (2023), 927--934. \url{https://doi.org/10.1093/glycob/cwad063}

\bibitem{adas2024}
Shengran Hu, Cong Lu, and Jeff Clune. 2025. Automated design of agentic systems. In \textit{Proceedings of the Twelfth International Conference on Learning Representations}. \url{https://arxiv.org/abs/2408.08435}

\bibitem{pid2024}
Shreeyash Gowaikar, Srinivasan Iyengar, Sameer Segal, and Shivkumar Kalyanaraman. 2025. An agentic approach to automatic creation of P\&ID diagrams from natural language descriptions. In \textit{AAAI Workshop on AI to Accelerate Science and Engineering}. \url{https://arxiv.org/abs/2412.12898}

\bibitem{datainterpreter2024}
Sirui Hong, Yizhang Lin, Bang Liu, Bangbang Liu, Binhao Wu, Ceyao Zhang, Chenxing Wei, Danyang Li, Jiaqi Chen, Jiayi Zhang, Jinlin Wang, Li Zhang, Lingyao Zhang, Min Yang, Mingchen Zhuge, Taicheng Guo, Tuo Zhou, Wei Tao, Xiangru Tang, Xiangtao Lu, Xiawu Zheng, Xinbing Liang, Yaying Fei, Yuheng Cheng, Zhibin Gou, Zongze Xu, and Chenglin Wu. 2024. Data Interpreter: An LLM agent for data science. \textit{arXiv:2402.18679}. \url{https://arxiv.org/abs/2402.18679}

\bibitem{polymind2025}
Qian Wan, Jiannan Li, Huanchen Wang, and Zhicong Lu. 2025. Polymind: Parallel visual diagramming with large language models to support prewriting through microtasks. \textit{arXiv:2502.09577}. \url{https://arxiv.org/abs/2502.09577}

\bibitem{mathagents2023}
Melanie Swan, Takashi Kido, Eric Roland, and Renato P. dos Santos. 2023. Math agents: Computational infrastructure, mathematical embedding, and genomics. \textit{arXiv:2307.02502}. \url{https://arxiv.org/abs/2307.02502}

\end{thebibliography}
\end{document}